\documentclass[11pt]{article}

\usepackage[top=1in, bottom=1in, left=1in, right=1in]{geometry}
\usepackage{amsmath,amssymb,epsfig}  
\usepackage{epstopdf}
\usepackage{graphicx}  
\usepackage{subfig}
\usepackage{cite} 
\usepackage{array} 
\usepackage{enumerate}
\usepackage{multirow}
\usepackage{hhline}
\usepackage{color}
\usepackage{url}
\usepackage{authblk}

\title{Persistent Homology on Grassmann Manifolds for Analysis of Hyperspectral Movies}
\author[1]{Sofya Chepushtanova}
\author[2]{Michael Kirby} 
\author[2]{Chris Peterson}
\author[3]{Lori Ziegelmeier}

\affil[1]{Wilkes University, Wilkes-Barre, PA, USA \vskip.1ex \texttt{sofya.chepushtanova@wilkes.edu}}
\affil[2]{Colorado State University, Fort Collins, CO, USA \vskip.1ex \texttt{(kirby, peterson)@math.colostate.edu}}
\affil[3]{Macalester College, Saint Paul, MN, USA \vskip.1ex \texttt{lziegel1@macalester.edu}} 
         
\date{}    

\begin{document}
\maketitle
\begin{abstract}
The existence of characteristic structure, or shape, in complex data sets has been recognized as
increasingly important for mathematical data analysis. This realization has motivated the development of new tools such as persistent homology for exploring topological invariants, or features, in large data sets. In this paper we apply persistent homology to the characterization of gas plumes in time dependent sequences of hyperspectral cubes, \emph{i.e.} the analysis of 4-way arrays.  We investigate hyperspectral movies of Long-Wavelength Infrared data monitoring an experimental release of chemical simulant into the air. Our approach models regions of interest within the hyperspectral data cubes as points on the real Grassmann manifold $G(k, n)$ (whose points parameterize the $k$-dimensional subspaces of $\mathbb{R}^n$), contrasting our approach with the more standard framework in Euclidean space. An advantage of this approach is that it allows a sequence of time slices in a hyperspectral movie to be collapsed to a sequence of points in such a way that some of the key structure within and between the slices is encoded by the points on the Grassmann manifold. This motivates the search for topological features, associated with the evolution of the frames of a hyperspectral movie, within the corresponding points on the Grassmann manifold. The proposed mathematical model affords the processing of large data sets while retaining valuable discriminatory information. In this paper, we discuss how embedding our data in the Grassmann manifold, together with topological data analysis, captures dynamical events that occur as the chemical plume is released and evolves.
\end{abstract}

\section{Introduction}
\label{sec:intro}
Hyperspectral imaging (HSI) technology allows the acquisition of information across the electromagnetic spectrum that is invisible to humans.  In a very real sense, these cameras allow us to ``see the unseen" by including wavelengths spanning ultraviolet and far infrared. In contrast, humans can observe a very limited range of the electromagnetic spectrum, \emph{i.e.} wavelengths of approximately 400-700nm are visible to the human eye.  

\begin{figure}[htb]
\centering  \includegraphics[scale=.25]{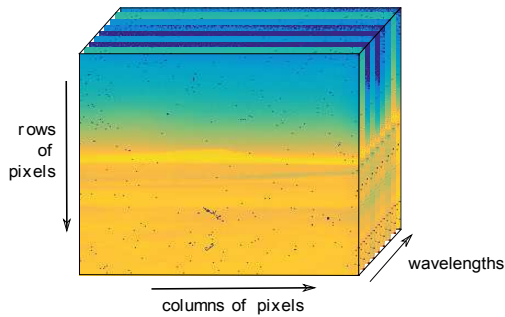}
 \caption{\small{Illustration of one frame, or data cube, of a hyperspectral movie collected with the Fabry-P\'erot Interferometer.}}
\label{fig_datacube}
\end{figure}

Multi- and hyper-spectral imaging technology has become widely available, and there is an increasing number of canonical data sets available for scientific analysis including, \emph{e.g.} the AVIRIS Indian Pines\footnote{Available from \url{https://engineering.purdue.edu/~biehl/MultiSpec.}} and the ROSIS University of Pavia\footnote{Available from \url{http://www.ehu.es/ccwintco/index.php/}} data sets.  In addition, moving objects may be detected with devices such as the Fabry-P\'erot Interferometer\cite{FPdata} which can capture {\it hyperspectral movies} at frame rates of approximately 5Hz. See Figure \ref{fig_datacube} for an illustration.  The resulting 4-way arrays of spatial-spectral-temporal data provide a high fidelity view of our environment and may help in the monitoring of pollution in the air and water.  An application that concerns us in this paper is the characterization of gaseous plumes as they are released into the environment.

Traditionally, one of the primary applications of hyperspectral image analysis consists of object detection and classification.  The focus is generally on the identification of anomalous pixels in the image and the determination of the composition of the materials in the pixel.  A range of mathematical tools have been developed for the analysis of hyperspectral images including, \emph{e.g.} matched subspaces, the RX algorithm, and the adaptive cosine estimator\cite{Manolakis2008}.  More recently, manifold learning algorithms have been applied to hyperspectral images to exploit topology and geometry, \emph{i.e.} mathematical shape, or signatures, in data at the pixel level \cite{bachmann2005exploiting,ma2010local}.

The subspace perspective is also taken in this paper,  but in the direction of understanding the topology and geometry of the Grassmann manifold (Grassmannian) associated with hyperspectral images, \emph{i.e.} the manifold parameterizing the $k$-dimensional subspaces of $n$-dimensional space.  While we are motivated by ideas similar to those found in prior applications of manifold learning algorithms, \emph{e.g.} \cite{bachmann2005exploiting,ma2010local}, our application data is not at the pixel level.  By constructing subspaces of pixels we are able to exploit the rich metric structure of the Grassmannian based on measuring angles between subspaces.  The advantage of this approach is that a set of pixels used to form a subspace is seen to capture the variability in the data missing in a single pixel observation.

An example that illustrates the power of this framework is the application to illumination spaces in the face recognition problem.  The variation in illumination on an object may be approximated by a cone captured in a low-dimensional subspace. Subspace angles can be used to compute similarity of illumination spaces and the effect on classification accuracy was striking when applied to  the CMU-PIE data set,  even on ultra-low resolution images \cite{chang}.  More recently, tools have been developed to represent points on Grassmannians via subspace means \cite{marrinan2014finding}, or nested flags of subspaces \cite{draper2014flag}. In another application to video sequence data, we used the setting of the Grassmannian to extend an algorithm on vector spaces for detection of anomalous activities \cite{WangID}. 

In this paper, we address the question of the existence of topological signatures in the setting of hyperspectral movies mapped to the Grassmannian.  Our approach builds on applying the Grassmannian architecture to hyperspectral movies that has shown promise in preliminary work \cite{Igarss,grasspaper}.  Here, our focus is on application of persistent homology (PH) to the characterization of the evolution of chemical plumes as acquired by hyperspectral movie data sets. 
As in the application to face recognition, we encode a single frame of a hyperspectral movie (or a collection of pixels of a single frame in the movie) as a point on the Grassmann manifold. 
We speculate that this manifold representation affords a form of compression of information while capturing essential topological structure. We consider the application of this approach to the characterization of chemical signals as measured by the Long-Wavelength Infrared (LWIR) data set \cite{FPdata}.   Our goal is to establish the existence of topological signatures that can provide insight into the evolution of complex 4-way data arrays.

The paper outline includes an overview of PH in Section~\ref{sec:persistence} and the geometry of the Grassmannian in Section~\ref{sec:grassmannian}. Computational experiments are discussed in Section~\ref{sec:results} and conclusions are given in Section~\ref{sec:conclusion}.

\section{Persistent Homology} 
\label{sec:persistence}

Homology is an invariant that measures features of a topological space and can be used to distinguish distinct spaces from one another \cite{Hatcher}. Persistent homology encodes a parameterized family of these homological features. It is a computational approach to topology that allows one to answer basic questions about the structure of point clouds in data sets at multiple scales \cite{carlsson2009topology}. This procedure involves (1) interpreting a point cloud as a noisy sampling of a topological space, (2) creating a global object by forming connections between proximate points based on a scale parameter, (3) determining the topological structure made by these connections, and (4) looking for structures that persist across different scales. PH has been used to understand the topological structure of data arising from applications including \cite{imagewebs, windowsandpersistence, hippocampalPH, corticalsurfacePD, visionTDA, swarms}.  For a detailed discussion of homology, see \cite{Hatcher}, and for further discussions of persistent homology, see \cite{barcodes, edelsbrunner2008persistent,carlsson2009topology}.

One way to associate a family of topological objects with a point cloud is to use the points to construct a family of nested simplicial complexes. The \emph{Vietoris-Rips} complex builds a simplicial complex $S_\epsilon$ with vertices as the data points and higher dimensional $k$-simplices formed whenever $k+1$ points have pairwise distances less than $\epsilon$. The $k$-dimensional holes of this simplicial complex generate a homology group $H_k(S_\epsilon)$ whose rank, known as the $k$-th Betti number, counts the number of $k$-dimensional holes. For instance, $Betti_0$ measures  the number of connected components (clusters) of the point cloud, while $Betti_1$ indicates the existence of topological circles (loops), or periodic phenomenon. To avoid picking a specific scale $\epsilon$, persistent homology seeks structures that persist over a range of scales, exploiting the fact that as $\epsilon$ grows, so do the simplicial complexes $S_{\epsilon_1} \subseteq S_{\epsilon_2} \subseteq S_{\epsilon_3} \subseteq \ldots$ indexed by the parameters $\epsilon_1 \leq \epsilon_2 \leq \epsilon_3 \leq \ldots$. Thus, PH tracks homology classes of the point cloud along the scale parameter, indicating at which $\epsilon$ a $k$th order hole appears and for which range of $\epsilon$ values it persists. The Betti numbers, as functions of the scale $\epsilon$, can be visualized in a distinct \emph{barcode} for each dimension $k$ \cite{barcodes}.  
   
\begin{figure}[htb]
\begin{minipage}[b]{.48\linewidth}
  \centering
 \centerline{\epsfig{figure=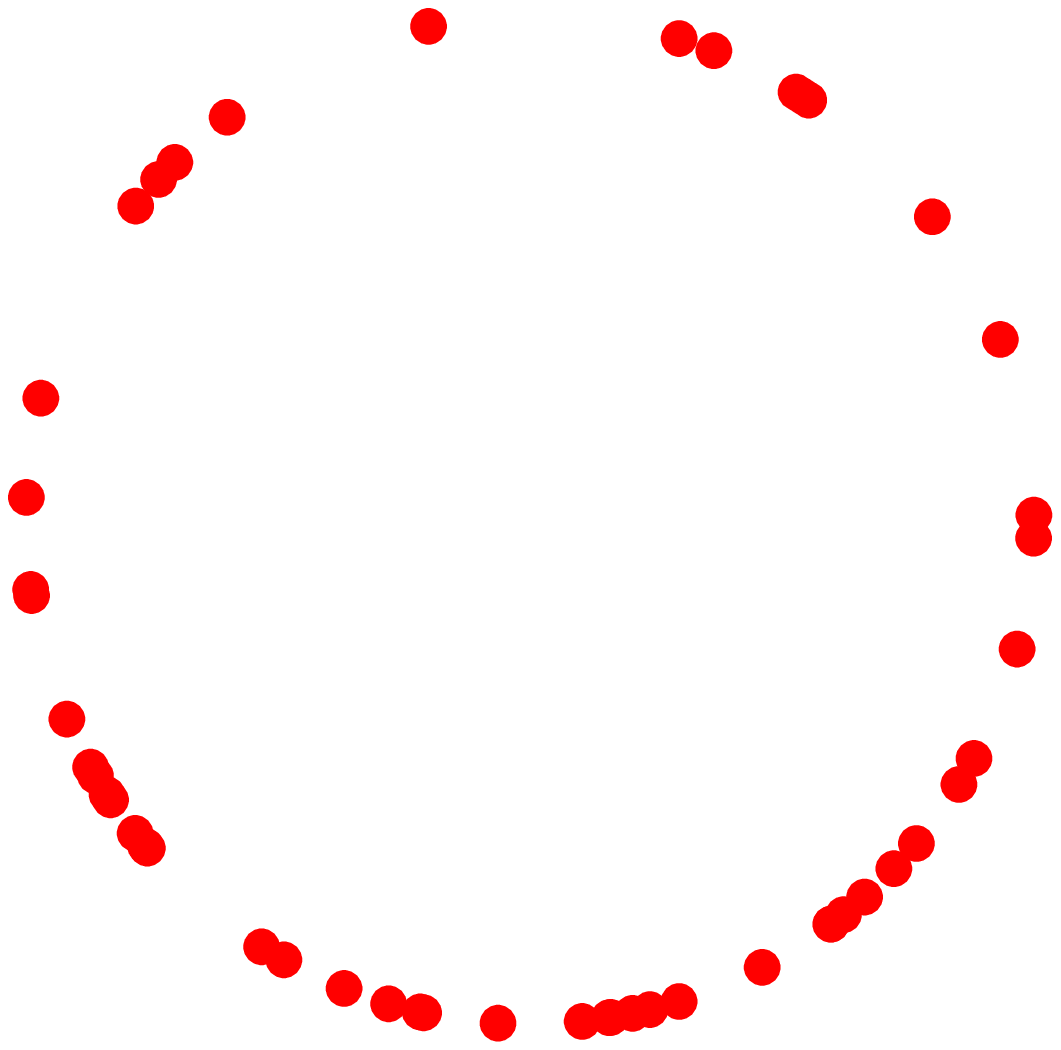,scale=0.3}}
\medskip
\end{minipage}
\hskip-1ex 
\begin{minipage}[b]{0.52\linewidth} 
  \centering
 \centerline{\epsfig{figure=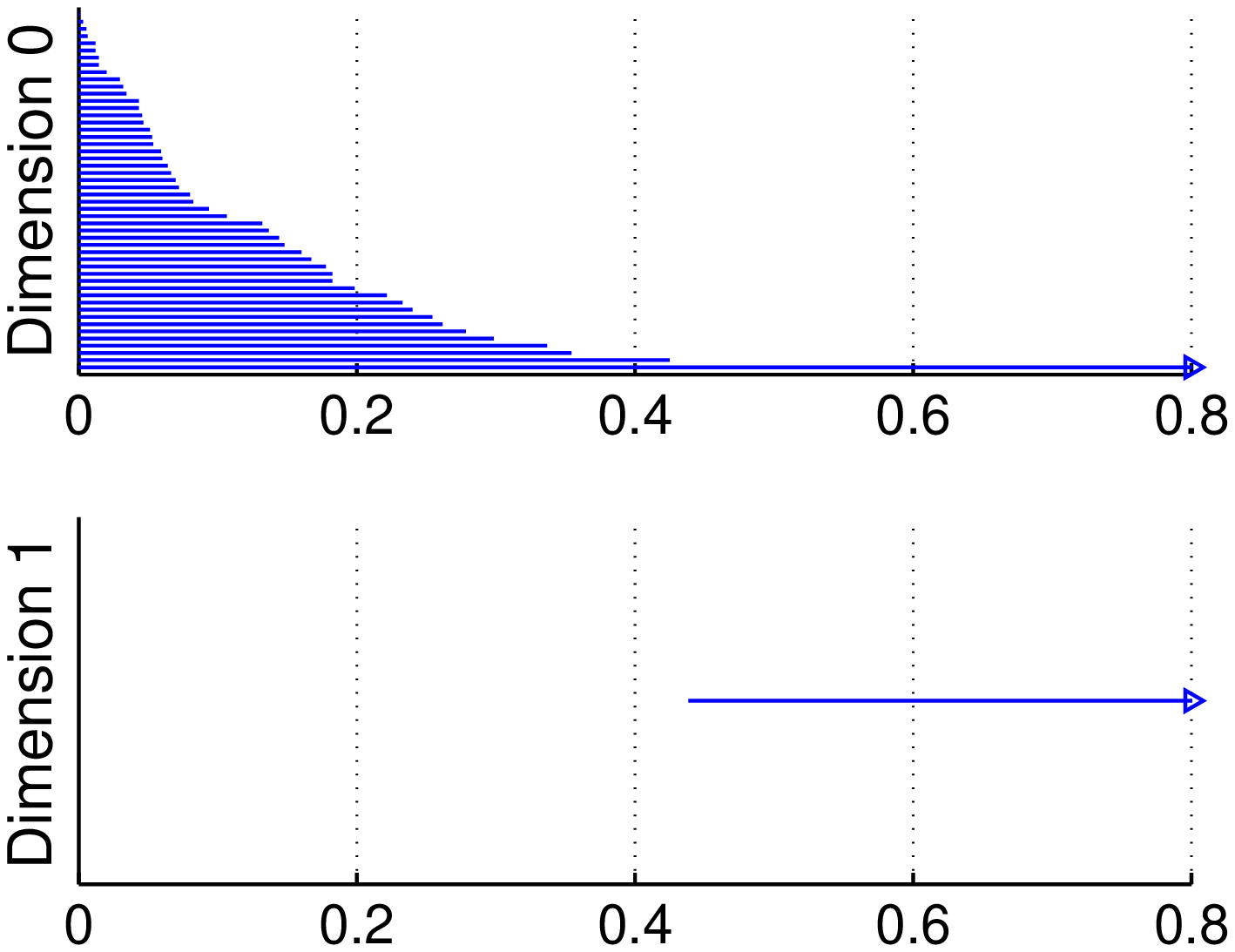,scale=0.3}}
\medskip
\end{minipage}
\caption{\small{$Betti_0$ (top right) and $Betti_1$ (bottom right) barcodes corresponding to point cloud data sampled from the unit circle (left).}}
\label{fig_circle}
\end{figure}

Figure~\ref{fig_circle} is an example of the $k=0$ and $k=1$ barcodes generated for a point cloud sampled from a circle. Each horizontal bar begins at the scale where a topological feature first appears and ends at the scale where the feature no longer remains. The $k$th Betti number at any given parameter value $\epsilon$ is the number of bars that intersect the vertical line through $\epsilon$. Short-lived features are often considered as noise while those features persisting over a large range of scale represent true topological characteristics. In the case of $Betti_0$, at small values of $\epsilon$ there will be a distinct bar for each point, as the simplicial complex $S_\epsilon$ consists of isolated vertices. At large values of $\epsilon$, only one bar remains, as all data will eventually connect into a single component. For the circle, $Betti_0 = Betti_1 = 1$ which correspond to the number of connected components and number of loops, respectively, shown by the longest (persistent) horizontal bars in each plot. We use JavaPlex, a library for computing PH and TDA in this paper\cite{Javaplex}. 

\section{The Geometry of the Grassmann Manifold}
\label{sec:grassmannian}
The (real) Grassmann manifold $G(k,n)$ is a parameterization of all $k$-dimensional subspaces of $n$-dimensional space \cite{edelman}. 
A point on $G(k,n)$ can be represented by a tall $n \times k$ matrix $Y$ with the property
that $Y^TY= I_k$ where $Y$ is an element of the equivalence class $\lfloor Y \rfloor$ consisting
of all matrices of the form $YQ$ with $Q \in O(k)$, the orthogonal group that consists of $k \times k$ orthogonal matrices \cite{edelman}.  

Hyperspectral data is a 3-way cube $x \times y \times \lambda$  that can be mapped to points in a Grassmannian in a variety of ways.  Here, we select a subset of $k$ frequencies $\lambda_i$.  For each of the $k$ frequencies we propose
to ``vectorize" the $xy=n$ spatial components to form an $n \times k$ matrix $X$.  It is assumed that the construction is
such that $k < n/2-1$ so subspaces don't overlap trivially.  To map $X$ to a matrix $Y$ representing a point on
the Grassmann we compute any orthogonal basis for the column space of $X$. For instance, the $n\times k$ matrix $U$ in the {\it thin} singular value decomposition $X = U \Sigma V^T$ provides one option as a representation of a point on the Grassmanian $G(k,n)$.

The mapping described above allows us to construct a sequence of points on $G(k,n)$, each one taken from the same
spatial location in the 3-way array of hyperspectral pixels or from the same frame of a hyperspectral movie.  The pairwise distances between the points in this sequence are computed in terms of the principal angles between the subspaces.
The implementation of the Grassmannian framework is, in part, motivated by the rich metric structure of a variety of distance measures including the chordal, geodesic, and Fubini-Study distances,
which are all functions of the $k$ principal angles between the subspaces \cite{bjorck,conway}.

The experiments in this paper use the (pseudo)distance between two subspaces measured by the smallest principal angle. This (pseudo)distance has been effective in other numerical experiments \cite{chang, grasspaper}, and in fact,  we observed, in the experiments in this paper, that using it resulted in stronger topological signals than other distance measures.  
Once a distance matrix for the points on the Grassmannian is computed, we apply PH to determine topological structure.  In particular, we explore $Betti_0$ barcodes to estimate the number of connected components and $Betti_1$ barcodes to detect topological circles.  The goal is to associate physical properties in the HSI image that relate to these structures.

\section{Experimental Results}
\label{sec:results}
In this section, we apply PH to Long-Wavelength Infrared (LWIR) multispectral movies,  each of an explosive release of a chemical and resulting toxic plume which travels across the horizon of the scene \cite{FPdata}. The simulants released included Triethyl Phosphate (TEP) and Methyl Salicylate (MeS) in quantities of 75kg and 150kg, respectively. The LWIR data sets are captured using an interferometer in the 8-11 $\mu\text{m}$ range of the electromagnetic spectrum. A single {\it frame}, or data cube, of this movie consists of $256 \times 256$ pixels collected at $20$ IR bands. 
A given movie is a sequence of data cubes consisting of {\it pre-burst} and  {\it post-burst} frames.

The purpose of this paper is not to propose a new algorithm for detecting chemical plumes but rather to investigate the topological features associated with a known plume.  
The data processing workflow consists of the following steps: (1) band selection, (2) identification of the region containing the chemical plume, (3) mapping data  to the Grassmannian, (4) computing (pseudo)distances on the Grassmannian using the smallest principal angle, (5) determination of PH $Betti_0$ and $Betti_1$ barcodes, and, finally, (6) interpretation of the structure in the data as encoded by the topological invariants.  We describe more detail of steps (1) and (2) below.

\medskip

{\noindent \it Band Selection.} We applied the sparse support vector machine (SSVM) algorithm for optimal {\it in situ} band selection, \emph{i.e.} the SSVM identifies wavelengths that best discriminate the plume from the natural background \cite{chep}. 
In another approach, we visually choose bands which have the strongest plume signal in data cubes which have had the background removed and thus, have visible plume.

\medskip

\begin{figure}[pt!]
 \centering   
  \subfloat[]{\includegraphics[scale=0.31]{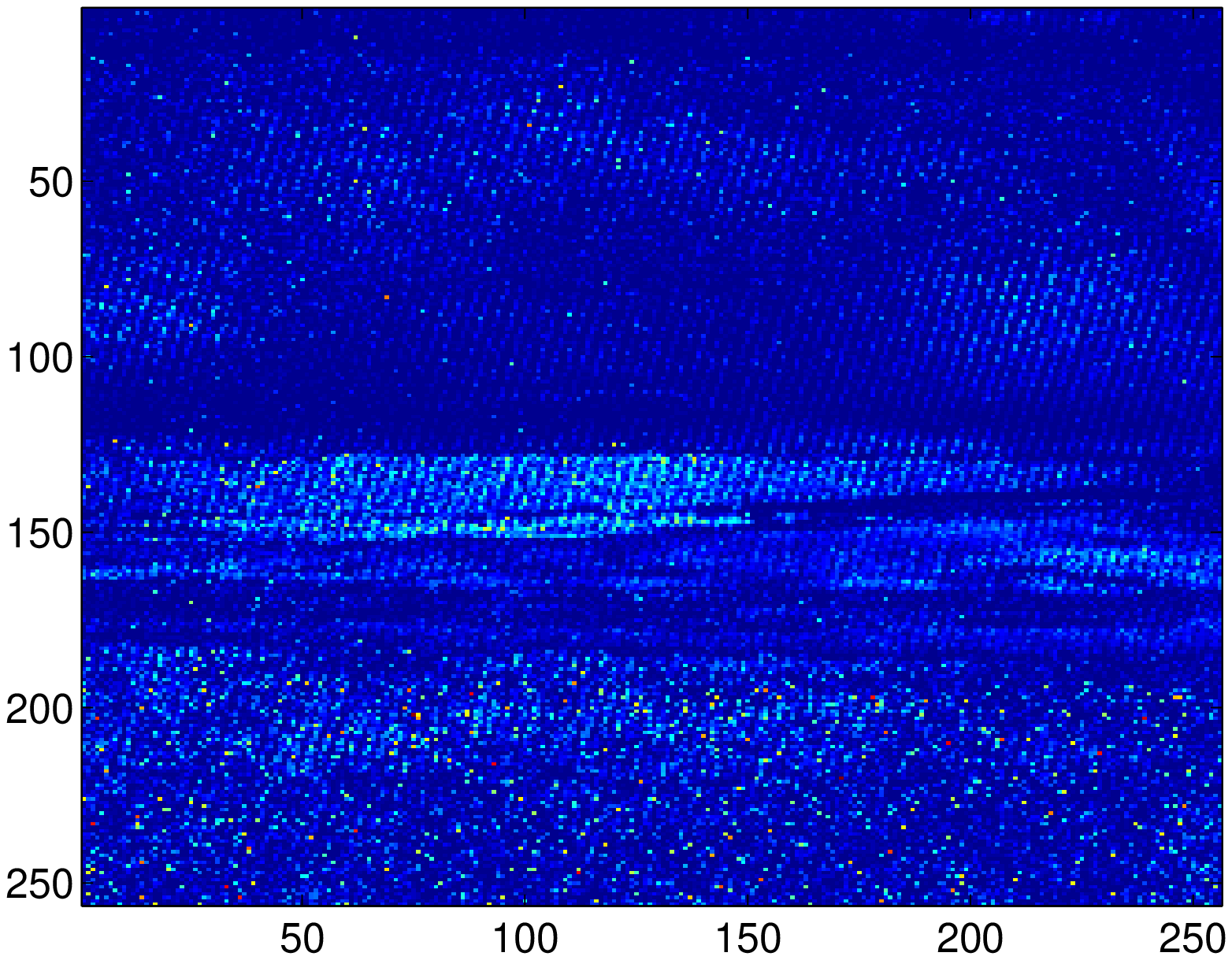}\label{fig_ACEforCubes_a}}.
  \subfloat[]{\includegraphics[scale=0.31]{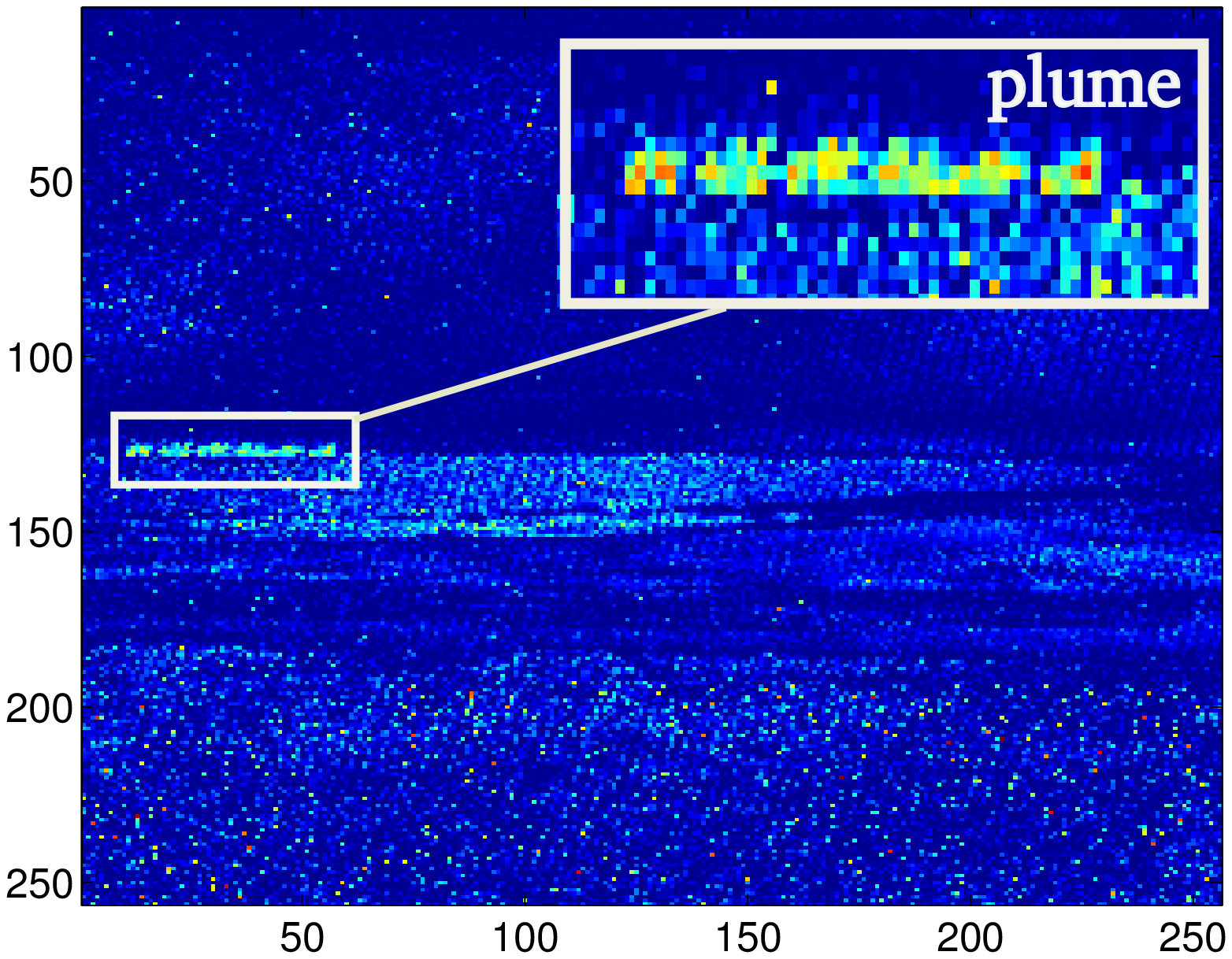}\label{fig_ACEforCubes_b}}.
  \caption[The ACE detector on the LWIR data cubes]
          {{\small{The ACE detector on LWIR data cubes: (a) ACE values of a pre-burst cube indicating that no plume is detected;
          (b) ACE values of a post-burst cube with a plume detected.  We have magnified the plume region to illustrate the performance of the ACE detector.}}}
 \label{fig_ACEforCubes}
\end{figure}

\noindent {\it Plume detection.}  The location of the chemical plume in the post-burst cubes is
determined using the well-known adaptive-cosine-estimator (ACE) \cite{Manolakis2008}. The ACE detector is one of the benchmark hyperspectral detection algorithms which computes the squared cosine of the angle between the whitened test pixel and the whitened target's spectral signature. Based on a chosen threshold, an ACE score indicates if the chemical is present in the test pixel. Figure~\ref{fig_ACEforCubes_a} depicts an image corresponding to a cube without a plume, and Figure~\ref{fig_ACEforCubes_b} depicts a cube with a chemical plume detected by the ACE.

\subsection{Experiment on Triethyl Phosphate Movie} 
\label{sec:allcubes}

We first consider the 561 frame multispectral movie of the data collection event of chemical Triethyl Phosphate (TEP) being released into the air. The data consists of the raw, unpreprocessed data including background clutter.  It was determined that the wavelengths $\{9.53, 8.30,10.68\}$ (nm) were optimal for discriminating TEP from background using the SSVM band selection algorithm. In this experiment we determine $Betti_0$ barcodes using all 561 TEP cubes, where $4 \times 8 \times 3$ subcubes have been extracted from regions of each data cube along the plume location region. 

\begin{figure}[htb]
\begin{center}
 \subfloat[]{\includegraphics[height=0.18\textheight,width=.50\textwidth]{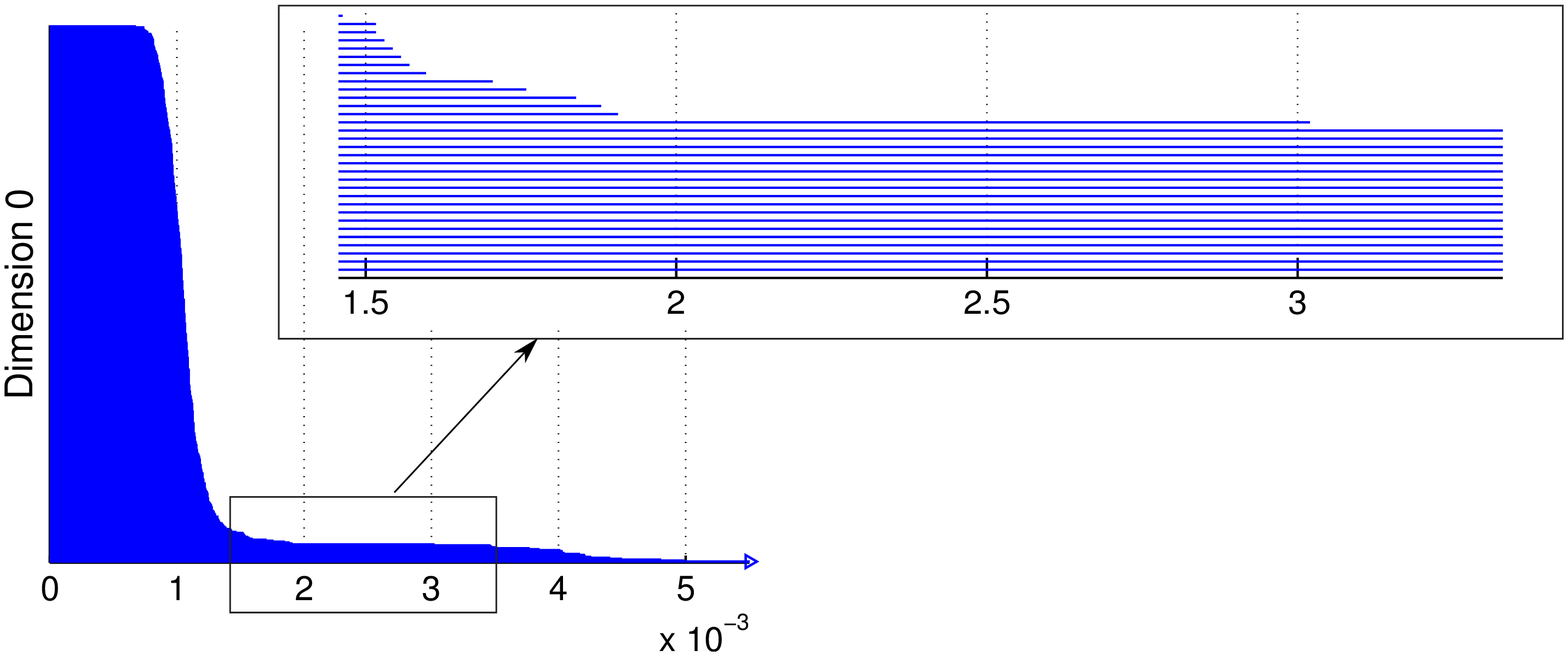}\label{fig_zoomedBarcode1}}
   \subfloat[]{\includegraphics[height=0.18\textheight,width=.50\textwidth]{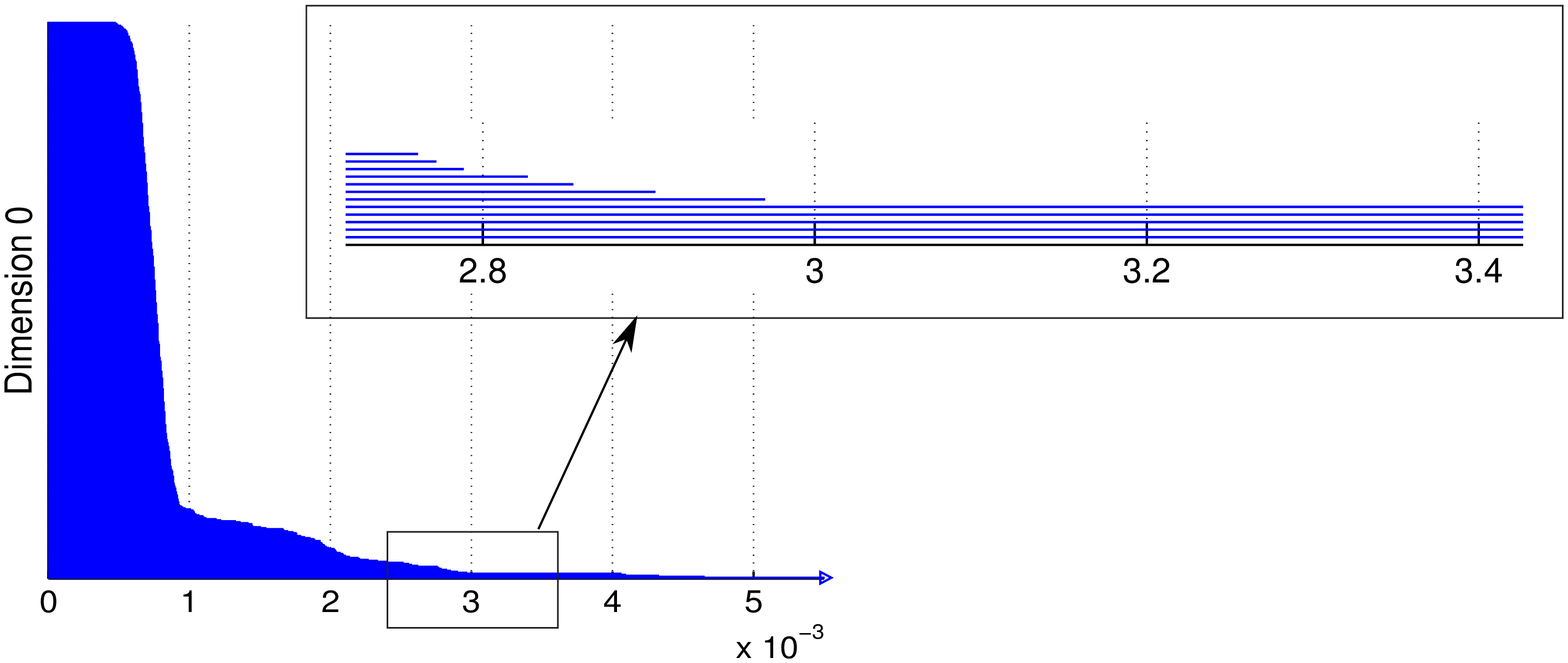}\label{fig_zoomedBarcode2}}
 \end{center}
\caption{{\small{(a) $Betti_0$ barcode generated on $4 \times 8 \times 3$ left horizon (plume formation) region
         limited by pixel rows 124-127 and columns 34-41,
         through all 561 TEP cubes, mapped to $G(3,32)$. (b) $Betti_0$ barcode generated on $4 \times 8 \times 3$ horizon region limited by pixel rows 124-127 and columns 75-82, through all 561 TEP cubes, mapped to $G(3,32)$.}}}
\label{fig_zoomedBarcodes}
\end{figure}

The $Betti_0$ barcode in Figure~\ref{fig_zoomedBarcode1} arises from the 561 Grassmanian points corresponding to the left horizon $4 \times 8 \times 3$ region in each data cube, limited by pixel rows 124 to 127 and pixel columns 34 to 41. This region belongs to the area when a plume forms and first becomes visible at frame 112 as detected by the ACE. At scale $\epsilon=1.5 \times 10^{-3}$, there are 31 bars corresponding to 31 connected components on $G(3,32)$, with 28 isolated points from frames 111 to 142, one cluster containing frames \{134, 135, 137\}, one cluster containing frame 519, and another containing all other frames.  At scale $\epsilon = 2 \times 10^{-3}$, we have 19 bars corresponding to 19 connected components on $G(3,32)$, with 18 isolated frames from 112 to 129, and one cluster containing all the rest. These bars persist for a large range of parameter value (to just beyond $3 \times 10^{-3}$), indicating a large degree of separation. At $\epsilon = 4 \times 10^{-3}$, we have 13 clusters with 11 isolated frames 112, 114 to 118, 120 to 123 and 125, one cluster of frames \{119, 124\}, and another containing everything else; see also \cite{Igarss}. Cubes following frame 111 are where the plume first occurs with the highest concentration of chemical, and the composition of the plume changes quickly as time progresses. PH detects separation of these cubes from pre-plume cubes and those cubes where plume no longer remains at multiple scales. 

After the plume is released, the plume drifts to the right in the multispectral movie as time progresses. We now consider a plume patch corresponding to a horizon region located to the right of the original plume location discussed above. That is, a $4 \times 8 \times 3$ patch is drawn from pixel rows 124 to 127 and pixel columns 75 to 82 for each of the 561 data cubes in the TEP movie. This data is embedded in $G(3,32)$, and PH is implemented to uncover the structure of the data. Figure~\ref{fig_zoomedBarcode2} contains the $0$-dimensional barcode. Analyzing connected components as $\epsilon$ varies, we observe that they differ from those found in the previous experiment, see Figure~\ref{fig_zoomedBarcode1}. At scale $\epsilon=1.5 \times 10^{-3}$, we have 52 connected components on $G(3,32)$ corresponding to 47 isolated points from 119 to 141, 145 to 165, and 170 to 172. The other points are connected into four smaller clusters \{142,143,144\}, \{166,167\}, \{168,169\}, and \{173,174\}, and one  cluster containing all the other points. At scale $\epsilon=2 \times 10^{-3}$, there are 30 connected components on the Grassmannian, including 25 isolated points from 119 to 127, 129 to 140, 149, and 151 to 156; four clusters each containing \{128,136-138\}, \{141-150\}, \{157,158\}, \{162-164\}; and one cluster containing all the rest. Further, at scale $\epsilon = 3 \times 10^{-3}$, the barcode plot has 5 bars that persist over a large range of values, namely, up to a little beyond $4 \times 10^{-3}$: 4 isolated points from frame 121 to 124 and one cluster containing all the rest.

We observe that for this region, PH separates points from frame 119 and later, in contrast to the frames separated from frame 112 in the previous experiment (Figure~\ref{fig_zoomedBarcode1}).
Note that points corresponding to frames 112 to 118 are ``plume-free'' as the plume does not reach this region until frame 119. It is also interesting to note that points corresponding to frames 121 to 124 are kept isolated for a large range of scales, \emph{i.e.} they are far away from each other and the rest of the points. PH, under the Grassmannian framework, treats these frames as being the most distinct in this region.

\subsection{Experiments Detecting A Loop in Methyl Salicylate Movie}

The next two experiments consider the multispectral movie of the data collection event of chemical Methyl Salicylate (MeS) being released into the air, consisting of 829 frames. Here we use 3 out of 20 wavelength bands \{10.57,10.68,10.94\} (nm) that were determined by visual inspection of a background-removed data cube where plume was present. These bands, in particular,  were selected as strong plume signal was visible at these corresponding wavelengths. In this movie, the plume first becomes visible at frame 32. 

In the first experiment, we construct a sliding window along the horizon, where the plume is released, in both a frame with and without a plume present (frames 32 and 1, respectively) to compare the topological structure of each. This sliding window is constructed by selecting $4\times 8 \times 3$ patches of each frame limited by rows 125-128 and columns 190-245 where each new point samples 8 columns, incrementing by one. Each patch is then embedded into $G(3,32)$ and the topological structure is analyzed with PH. In this experiment and the next, our focus is on the $Betti_1$ information which measures the number of loops present in the data. 
\begin{figure}[htb]
\begin{center}
\subfloat[]{\includegraphics[width=.4\textwidth,height=.22\textheight]{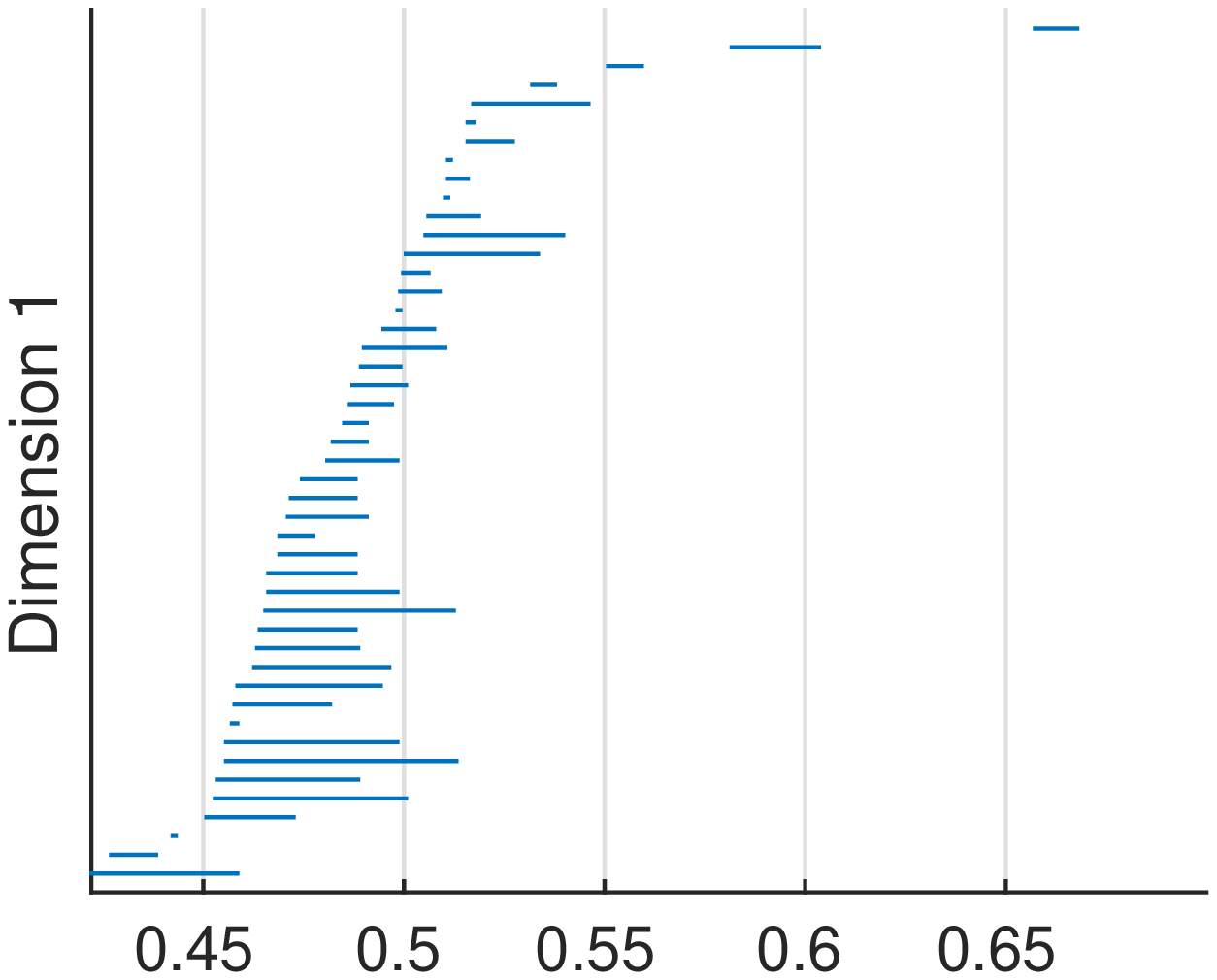}\label{fig:MeS150_frame1}} 
\subfloat[]{\includegraphics[width=.4\textwidth,height=.22\textheight]{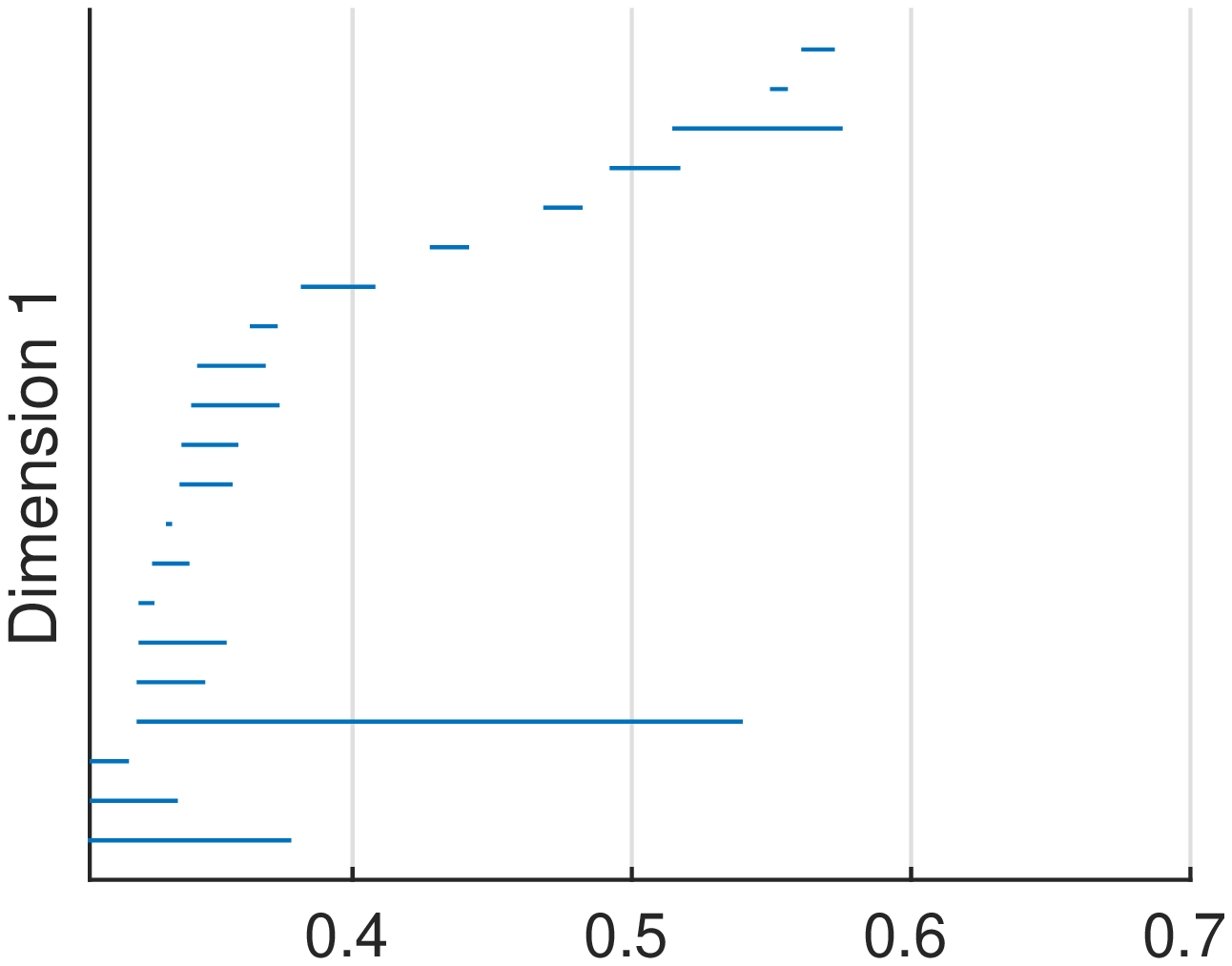}\label{fig:MeS150_frame32}} \hfill
 \end{center}
\caption{{\small{Data constructed by sliding a window along the horizon region of a single frame of the MeS movie, embedded into $G(3,32)$ and analyzed with PH. (a) $Betti_1$ barcode of frame 1. (b) $Betti_1$ barcode of frame 32. Observe that a persistent loop is present.}}}
\label{fig:slidingwindow}
\end{figure}

Observe in Figure \ref{fig:slidingwindow} that no persistent topological circle is present in the $Betti_1$ barcode of frame 1, while a persistent loop is present in the $Betti_1$ barcode of frame 32. This is interpreted as follows. In frame 32, where a plume is present, the sliding window first constructs points in $G(3,32)$ of the natural background, then traverses through points that contain plume, finally returning to points of the natural background. This creates a closed loop in $G(3,32)$. This behavior is captured in the topological structure of the plume cube. On the other hand, the sliding window in frame 1 only has points in $G(3,32)$ of the natural background, and thus, no persistent loop is formed in this space.

We mention that this experiment was done on background removed frames. In analysis with raw data, loops were not as prevalent with this framework. However, the next experiment does in fact use raw data in our analysis.

In the second experiment, we consider the first one hundred frames of the MeS cubes and focus on a ``plume location'' patch of size $4\times 8 \times 3$, limited by pixel rows 125 to 128 and pixel columns 217 to 224, embedded into $G(3,32)$ for each cube.

Figure \ref{fig:MeS150barcodes} displays the $Betti_0$ and $Betti_1$ barcodes from applying PH to this Grassmannian data. A fairly persistent bar appears in the $Betti_1$ barcode that begins at $\epsilon=0.00979$ and ends at $\epsilon=0.0141$.  This represents a loop through the data in $G(3,32)$. All other bars are considered as noise. Let us inspect this loop further. It begins once all of the data has been connected into a single component (refer to $\epsilon=0.00979$ in the $Betti_0$ barcode). The maximum pairwise distance--measured by the smallest angle between subspaces--for this data is 0.0308. This loop persists until just under half this distance. 

\begin{figure}[htb]
\begin{center}
\subfloat[]{\includegraphics[width=.6\textwidth,height=.2\textheight]{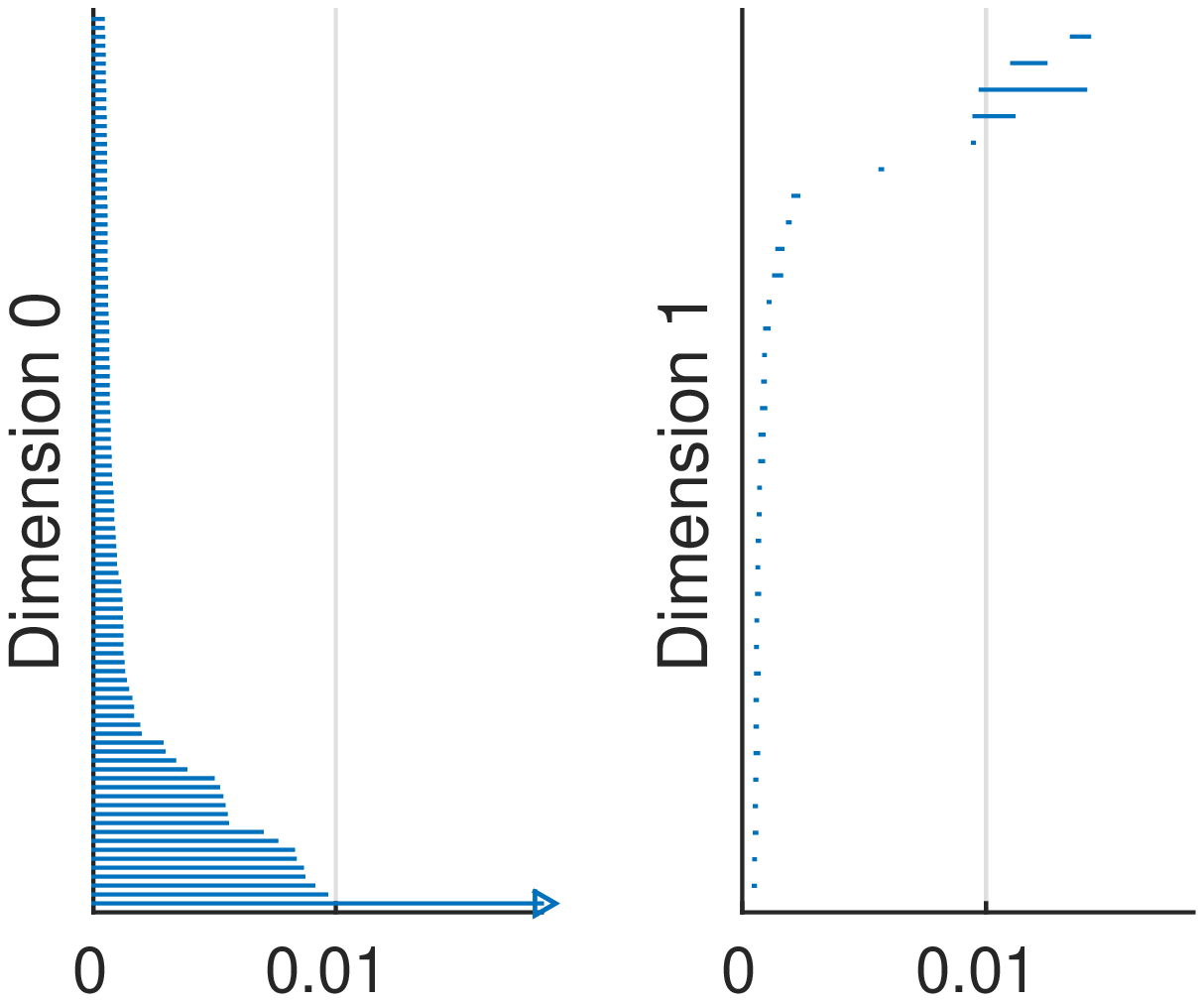}\label{fig:MeS150barcodes}} 
   \subfloat[]{\includegraphics[height=.2\textheight]{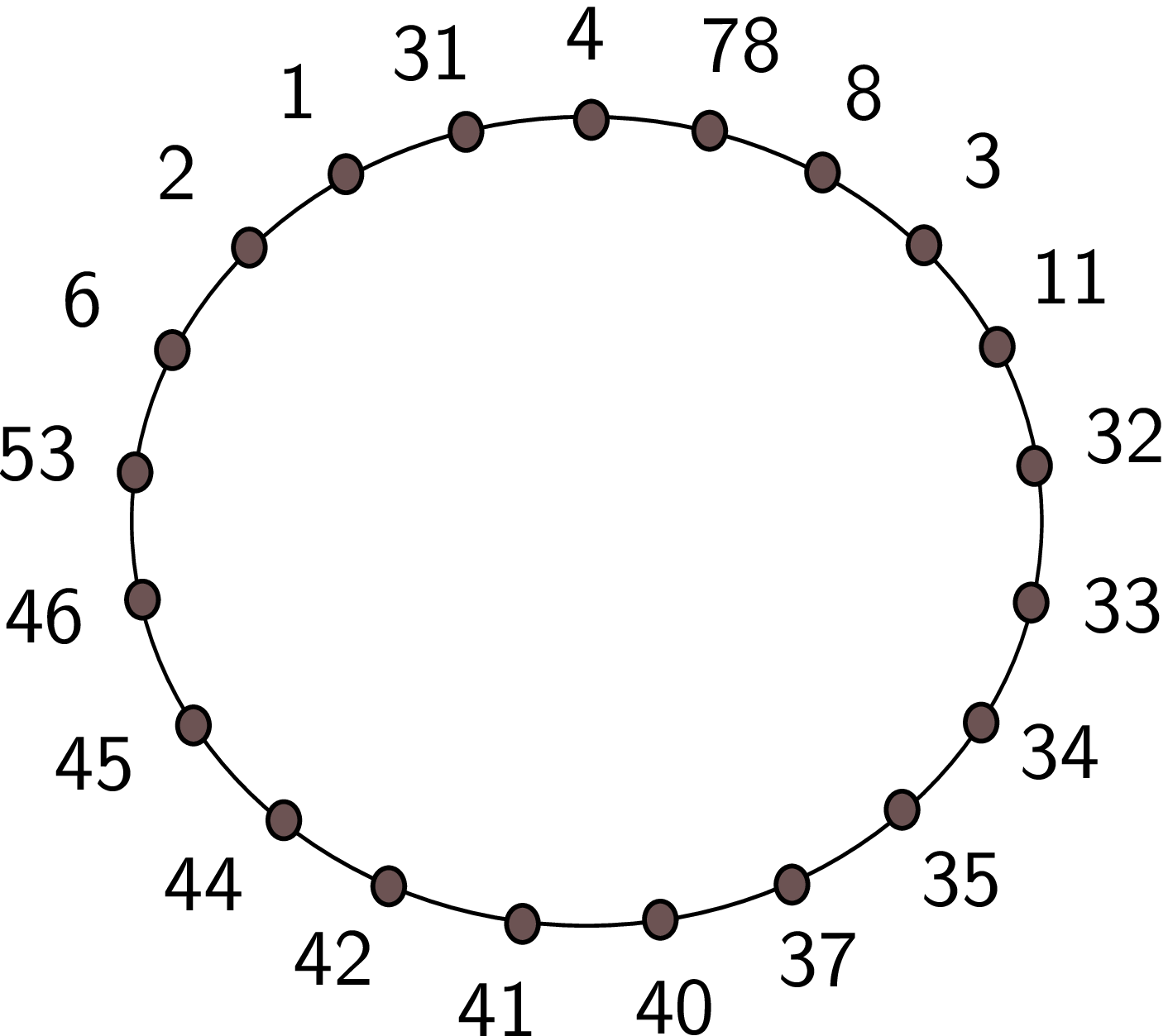}\label{fig:loop}} \hfill
 \end{center}
\caption{{\small{(a) $Betti_0$ and $Betti_1$ barcodes generated on $4 \times 8 \times 3$ plume locations limited by pixel rows 125-128 and columns 217-224, of the first one hundred frames of the MeS movie, mapped to $G(3,32)$. (b) Schematic of the edges forming the persistent $Betti_1$ feature.}}}
\label{fig:MeS150barcodes}
\end{figure}

We conclude the following from this experiment. The first few frames start with a fixed background. Then, the plume begins to form, spreading through the plume patch until the plume no longer remains in the $4 \times 8 \times 3$ sampled region. The remaining cubes then return to a fixed background, reflecting periodic behavior in the data. This collection of cubes traces out a closed loop, encoded in the Grassmann manifold $G(3,32)$. PH captures this loop in the persistent $Betti_1$ bar.  Figure \ref{fig:loop} displays a schematic of one possibility in the equivalence class of the edges that form this loop. While not all data cubes are present, we notice that those cubes immediately following 31 connect to one another sequentially. This is when the chemical is first released and begins to evolve. Cubes before this frame (where no plume is present) do not follow sequentially and connect with later cubes which no longer contain plume in the sampled plume patch. That is, the time dependent information of `pre-plume' and `post-plume' cubes--which simply contain information about the natural background and not the evolving plume--is not as important as `plume' cubes. 

\section{Conclusion}
\label{sec:conclusion}

We propose a geometric and topological model for capturing dynamical changes in hyperspectral movies. The HSI data cubes (or a sequence of pixel patches) are viewed as a sequence of points on the Grassmann manifold.  The tools of persistent homology are then applied to capture topological novelty in the setting of the Grassmann manifold.  This approach models cubes as points, a technique that permits the processing of potentially large amounts of data while retaining basic dynamical structure.
 
The dynamic structure recorded by the multispectral movie of the gas plume consisting of the simulant Triethyl Phosphate was illuminated in the $Betti_0$ barcodes.  Frames containing the plume were identified as topological singletons, \emph{i.e.} isolated points on the manifold for large ranges of scale.  Grassmann points before the release, as well as long after the release, appeared as clusters of points.  At a location to the right of this region, we see that later frames had a similar behavior, indicating that the geometric model of the Grassmannian allows the dynamics of the scene to be effectively characterized in a topological sense.  

In the next two experiments, we use the $Betti_1$ barcode on the movie of the release of Methyl Salicylate mapped to the Grassmannian to reveal that a closed loop is present on the manifold, again reflecting the evolution of the plume. First, we consider a sliding window of pixels along the plume location region and observe that a loop is present in a frame with a plume unlike a frame without a plume. Second, we consider a patch of pixels in each of the first one hundred frames and observe a closed loop that encompasses frames immediately following the release of the chemical in a sequential manner. We mention that in other HSI movies in the LWIR data set, when the amount of chemical released was not as much as in the MeS cubes, the signal of this loop was not as strong. These experiments illustrate that the use of the Grassmann manifold together with PH provide insight into the presence and concentration of chemical contamination in a hyperspectral movie.

\section*{Acknowledgments}

This paper is based on research partially supported by the National Science Foundation grants DMS-1228308, DMS-1322508.
Any opinions, findings, and conclusions or recommendations expressed in this material are
those of the authors and do not necessarily reflect the views of the National Science Foundation. 


\bibliographystyle{plain}
\bibliography{CTIC2016references}

\end{document}